\documentclass[journal]{IEEEtran}
\usepackage{cite}
\usepackage{amsmath}
\usepackage{graphicx}
\ifCLASSINFOpdf

\else

\fi

\begin{document}

\title{Rapidly and accurately estimating brain strain and strain rate across head impact types with transfer learning and data fusion}

\author{Xianghao~Zhan,
		Yuzhe~Liu,
		Nicholas~J.~Cecchi,
		Olivier~Gevaert,
		Michael~M.~Zeineh,
		Gerald~A.~Grant,
		and~David~B.~Camarillo
\thanks{This work was supported by the Department of Bioengineering, Stanford University and the Office of Naval Research Young Investigator Program (N00014-16-1-2949).}
\thanks{X. Zhan, Y. Liu, N. Cecchi, and D. Camarillo are with the Department of Bioengineering, Stanford University, Stanford, 94305, USA. (Corresponding Author: Y. Liu e-mail: yuzheliu@stanford.edu)}% <-this % stops a space
\thanks{O. Gevaert is with the Department of Biomedical Data Science and Stanford Center for Biomedical Informatics Research, Stanford University, Stanford, 94305, USA}% <-this % stops a space
\thanks{M. Zeineh is with the Department of Radiology, Stanford University, Stanford, 94305, USA}
\thanks{G. Grant is with the Department of Neurosurgery, Stanford University, Stanford, 94305, USA}% <-this % stops a space
}

\markboth{}%
{Shell \MakeLowercase{\textit{et al.}}: Bare Demo of IEEEtran.cls for IEEE Journals}

\maketitle

\begin{abstract}
Brain strain and strain rate are effective in predicting traumatic brain injury (TBI) caused by head impacts. However, state-of-the-art finite element modeling (FEM) demands considerable computational time in the computation, limiting its application in real-time TBI risk monitoring. To accelerate, machine learning head models (MLHMs) were developed, and the model accuracy was found to decrease when the training/test datasets were from different head impacts types. However, the size of dataset for specific impact types may not be enough for model training. To address the computational cost of FEM, the limited strain rate prediction, and the generalizability of MLHMs to on-field datasets, we propose data fusion and transfer learning to develop a series of MLHMs to predict the maximum principal strain (MPS) and maximum principal strain rate (MPSR). We trained and tested the MLHMs on 13,623 head impacts from simulations, American football, mixed martial arts, car crash, and compared against the models trained on only simulations or only on-field impacts. 
The MLHMs developed with transfer learning are significantly more accurate in estimating MPS and MPSR than other models, with a mean absolute error (MAE) smaller than $0.03$ in predicting MPS and smaller than $7 s^{-1}$ in predicting MPSR on all impact datasets. The MLHMs can be applied to various head impact types for rapidly and accurately calculating brain strain and strain rate. 
Besides the clinical applications in real-time brain strain and strain rate monitoring, this model helps researchers estimate the brain strain and strain rate caused by head impacts more efficiently than FEM.

\end{abstract}


\IEEEpeerreviewmaketitle

\section{Introduction}
%1. TBI & mTBI & monitoring
Traumatic brain injury (TBI) is a global health hazard that affects more than 55 million people worldwide and contributes to more than 2.2 million emergency department visits and more than 200 thousand hospitalizations in the United States \cite{james2019global}. TBI can be caused by head impacts from various sources including traffic accidents, blasts, various contact sports and domestic abuse \cite{caswell2017characterizing,cecchi2019head,corrigan2003early,hernandez2015six,o2020dynamic,versace1971review,wilcox2014head}. Additionally, mild TBI, instead of directly causing death or disabilities, can be harder to detect and the detection is important because TBI may lead to accumulation of brain damage over time, which can progress to long-term cognitive deficits and even neurodegeneration \cite{doherty2016blood}. The prevalence and the severe consequences of TBI patients highlight the need for better monitoring of brain injury risks because early detection and intervention are essential for recovery and prevention of further brain damage \cite{guiza2017early}. 

%2. Brain strain, strain rate (Susan Marguelles) & FEM & BIC & limited in time or feature dimensionality
Brain strain, particularly maximum principal strain (MPS), has been regarded as an effective predictor of TBI \cite{o2020dynamic,gabler2019development,gabler2018development}. Although a comprehensive understanding of the biological consequences resulting from mechanical loadings of the head to the resultant brain injury are not fully understood, strain-based metrics correlate well with brain injury \cite{bain2000tissue, bar2016strain, donat2021biomechanics, cater2006temporal,fahlstedt2015correlation}. In addition to brain strain, the maximum principal strain rate (MPSR) correlates well with traumatic axonal injury (TAI) \cite{hajiaghamemar2020head,hajiaghamemar2020embedded}, which suggests that TBI is dependent on not only the brain strain but also the strain rate. To compute the brain strain and strain rate, the head kinematics, which can be measured by various wearable devices, are used as the input \cite{hernandez2015six,camarillo2013instrumented,greenwald2008head}. Finite element modeling (FEM) is the state-of-the-art method for the computation of brain strain and strain rate \cite{takhounts2003development,kleiven2007predictors,mao2013development}. The major limitations of FEM are lengthy computational time and complicated computational software, which may not be easily accessible to researchers, clinicians and sport teams \cite{zhan2021rapid}. These limitations prevent the brain strain and strain rate from becoming widely used as the metrics in quantifying TBI risks in real-world applications because there may be hours passed after an American football players got impacted when the TBI risks are finally available. To complement the FEM, brain injury criteria have been developed by researchers with reduced-order mechanical models and statistical data fitting \cite{laksari2020multi,takhounts2013development,kimpara2012mild,zhan2021relationship,versace1971review}. These brain injury criteria can be calculated quickly based on the head kinematics. However, most brain injury criteria only output a single value, which limits the region-specificity of the risk information and makes it hard to directly interpret the TBI risks from the perspective of brain biomechanics. Furthermore, the effectiveness of brain injury criteria is generally tested on the prediction of brain strain rather than the strain rate \cite{laksari2020multi,takhounts2013development,zhan2021relationship}, and those brain injury criteria highly correlated with MPS are regarded as more accurate TBI risk estimators.

%3. MLHM: MLHM1.0, Ji*2: generalizability (accuracy decrease): strategy: type-specific modeling (MPSC1 and MPSC2): hard to realize when small dataset+high dimensional output, dataset diversity (inclusiveness), brain strain -> brain strain rate
Newly developed machine learning head models (MLHMs), which can be regarded as function approximators of FEM, start to address the issues of computational cost of FEM and the information-dimensionality limitation intrinsic to brain injury criteria 
\cite{zhan2021rapid,ghazi2021instantaneous}. Once the MLHMs are trained, the brain strain can be computed within a second, which may be helpful for near real-time TBI risk monitoring. However, the previously developed MLHMs are limited in several ways. First, they only model brain strain rather than brain strain and strain rate. To get strain rate, one still needs to resort to the complicated and time-consuming FEM. Second, the generalizability of these MLHMs is an issue significantly impacting its real-world applications: the model accuracy is not guaranteed across different types of head impacts, which is also a typical problem for the conventional risk estimation models like brain injury criteria \cite{zhan2021relationship}. This second limitation is more evident on small on-field impact datasets (not simulated impacts but real-world impacts recorded with sensors), because there are data distribution drifts from the simulated data (on which the MLHMs are trained) to the specific types of on-field data. However, there may not be enough on-field impacts collected to fully manifest the on-field data distributions to enable accurate predictions. For instance, our MLHMs showed higher accuracy on the head model simulated impacts and the college football (CF) impacts, but significantly lower accuracy on the mixed martial arts (MMA) impacts \cite{zhan2021rapid}. These two limitations leave the MLHMs as neither robust nor comprehensive enough to guarantee accurate estimation of TBI risks caused by various on-field head impacts.

%4. Data fusion & Transfer learning with the relatively easy-to-get simulated data as the basis.
To improve the generalizability across different types of head impacts, particularly the TBI risk estimation accuracy on small on-field impact datasets, and further optimize the MLHMs to be both fast and accurate to predict the strain rate from the perspective of model optimization, we developed two strategies for more powerful MLHMs: data fusion and transfer learning. With these two strategies, we developed a series of MLHMs on the impact datasets from simulation, CF, MMA, and car crashes to be able to predict both MPS and MPSR rapidly and accurately for specific types of head impacts. The abbreviations used in this study are shown in Table 1.

\begin{table}
		\centering
		\caption{The major abbreviations used in this study.}
		\label{tab:2}       % Give a unique label
		\begin{tabular}{cc}
			\hline\noalign{\smallskip}
			Abbreviation & Meaning \\
			\noalign{\smallskip}\hline\noalign{\smallskip}
			TBI & traumatic brain injury \\
			FEM & finite element modeling \\
			MLHM & machine learning head model \\
			MPS & maximum principal strain \\
			MPSR & maximum principal strain rate \\
			MAE & mean absolute error \\
			TAI & traumatic axonal injury \\
			CF & college football \\
			MMA & mixed martial arts \\
			HM & head model \\
			NFL & National Football League \\
			NHTSA & National Highway Traffic Safety Administration \\
			BBB & blood-brain-barrier \\
			\hline\noalign{\smallskip}
		\end{tabular}
	\end{table}

\section{Methods}
\subsection{Data description}
To prepare a dataset consisting of a variety of head impacts from different sources, we collected the head kinematics from a total of 13,623 head impacts from various sources: 1) 12,780 laboratory-reconstructed head impacts (labeled as HM for head model) simulated from a validated finite element (FE) model of the Hybrid III anthropomorphic test device (ATD) headform \cite{giudice2019development}; 2) 302 college football (labeled as CF) head impacts which were recorded by the Stanford instrumented mouthguard \cite{camarillo2013instrumented,liu2020validation}; 3) 457 MMA head impacts (labeled as MMA) which were recorded by the Stanford instrumented mouthguard \cite{tiernan2020concussion,o2020dynamic}; 4) 36 selected reconstructed helmeted head impacts from the National Football League  (labeled as NFL) \cite{sanchez2019reanalysis}; 5) 48 head impacts from automobile crashworthiness tests conducted by the National Highway Traffic Safety Administration (labeled as NHTSA). The distributions of the MPS, MPSR, peak resultant angular acceleration and peak resultant angular velocity are shown in Fig. 1.

\begin{figure*}[htbp]
    \centering
    \includegraphics[width=0.99\linewidth]{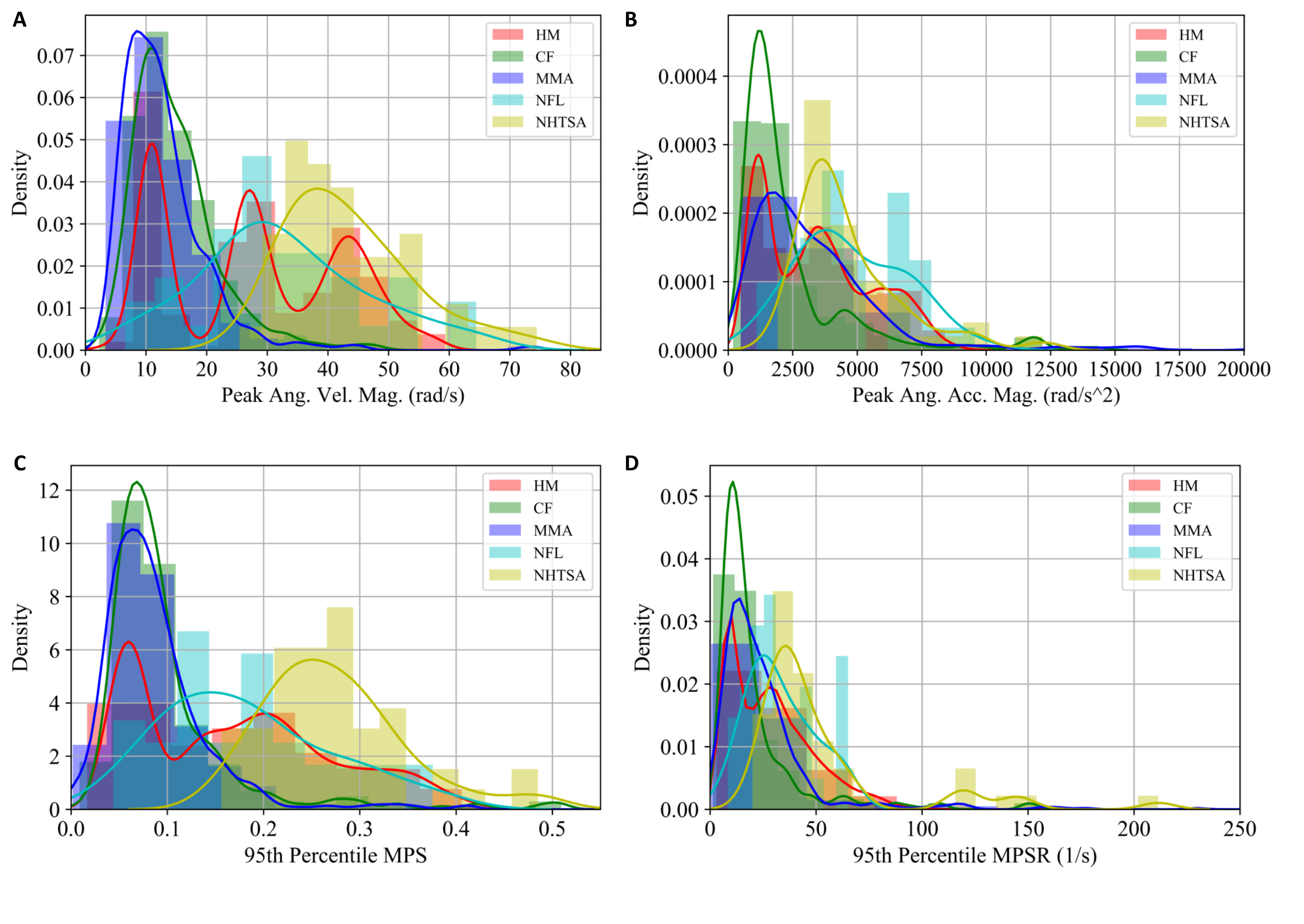}
    \caption{The distribution of the example kinematics features, MPS, and MPSR across various datasets. The distribution of the peak angular velocity (Ang. Vel.) magnitude (A), the peak angular acceleration (Ang. Acc.) magnitude (B), the 95th percentile MPS (C) and the 95th percentile MPSR (D).}
    \label{dist}
    \vspace{-5mm}
\end{figure*}

The head model simulated impacts were regarded as the basis dataset for model development before we performed data fusion between the on-field data and the basis dataset, or before we transferred the basis models trained on the simulated impacts to different on-field datasets. To improve the model accuracy, we simulated 12,780 impacts. First, the bare headform was impacted at different locations with velocities from 2 m/s to 8 m/s, which resulted in a total of 1,065 impacts \cite{zhan2021rapid}. The impact kinematics were processed via a second-order Butterworth low-pass filter (cut-off frequency: 150Hz). Then, considering the symmetry of the FE head model of the Hybrid III ATD, we performed the data augmentation by switching the kinematics along different axes (X: posterior-to-anterior, Y: left-to-right, Z: superior-to-inferior). This method was used to enlarge the impact datasets with 12 sets of impact with 1,065 impacts in each set. If XYZ is used to denote the kinematics of the 1,065 original impacts, we further enlarged the impact datasets with XNYZ, XZY, XZNY, YXZ, YZX, ZXY, NYZX, ZYX, ZXNY, ZNYX, NYXZ (the N before each axis denotes the negative kinematics along that axis). For example, the XZY denotes switching the original 1,065 impacts' kinematics between the Y axis and the Z axis. The on-field head impacts were published in previous studies, and the details of the post-processing can be found in \cite{zhan2021rapid,camarillo2013instrumented,liu2020validation,tiernan2020concussion,o2020dynamic,sanchez2019reanalysis}

The KTH FE model (Stockholm, Sweden) \cite{kleiven2007predictors}, which is a validated FE head model, was used to model the brain strain and strain rate caused by impacts. The model was developed in LS-DYNA (Livermore, CA, USA). The KTH FE model takes the brain, skull, scalp, meninges, falx, tentorium, subarachnoid cerebrospinal fluid (CSF), ventricles, and 11 pairs of the largest bridging veins in the modeling. Responses of the head model correlate well with experimental data of brain-skull relative motion \cite{kleiven2006evaluation}, intracranial pressure \cite{kleiven2006evaluation}, and brain strain \cite{zhou2019reanalysis}. In this study, to output more diverse and comprehensive risk information \cite{hajiaghamemar2020embedded,hajiaghamemar2020head} without losing the high-dimensional spatial information of the strain and the strain rate for each brain element, the MPS and MPSR for all 4,124 brain elements were extracted from the KTH model as the reference output values to be modeled by two series of deep learning head models (for MPS and MPSR, respectively). As for the computational cost, it usually takes 7 hours to model one impact with a computer (16GB RAM, Intel Core i7-6800K CPU).

\subsection{Head impact kinematics features}
Features, such as the peak value of angular acceleration magnitude, are important for brain strain prediction based on head impact kinematics, which have been shown in our previous study \cite{zhan2021rapid}. In this study, to get more accurate and generalizable brain strain and strain rate estimators, we extracted both temporal features and spectral features from the head kinematics by taking a union of the features extracted in two of our previous studies \cite{zhan2021rapid, zhan2021clustering}. The features were based on four different types of kinematics: 1) the linear acceleration at the head center of gravity $a(t)$, 2) the angular velocity $\omega(t)$, 3) the angular acceleration $\alpha(t)$, and 4) the angular jerk $j(t)$. The angular acceleration and the angular jerk were derived from the angular velocity via a five-point stencil numerical difference equation. For each type of kinematics, four channels were involved: three directions (X, Y, and Z) and the magnitude.

First, we extracted the 160 temporal features (based on angular velocity and angular acceleration) that were used in our previously developed MLHMs \cite{zhan2021rapid}, which included 20 features for each channel of signal: the maximum values, the minimum values, the integral values, the integral of absolute values, three sets of maximum and minimum values of the exponential moving average of the signal derivatives with different smoothing factors \cite{zhan2018discrimination,zhan2020electronic}, and the extrema information (the numbers of all positive extrema and of all negative extrema and the second to the fifth positive/negative extrema values). These features were extracted because they took both the signal intensities and time histories into consideration and proved to be effective in accurately predicting whole-brain strain, with specific details and reasons shown in our previous study \cite{zhan2021rapid}.

Second, to better improve the predictive power of the features and the generalizability of the models, we added another 350 temporal and spectral features that were used in another study \cite{zhan2021clustering}. The temporal features included the peaks of the absolute values of four channels for $a(t)$, $\omega(t)$, $\alpha(t)$, $j(t)$, which aggregated to 16 features. Furthermore, the square-root and squared features based on the 16 temporal features were included to account for the non-linearity in the relationship between kinematics and brain strain/strain rate. We have shown in our previous feature analysis study that the first-power, squared-root and second-power features were the most predictive for strain modeling (i.e., significantly more predictive than higher powers) \cite{zhan2021predictive}. Then, 304 spectral features were extracted \cite{zhan2021clustering}: we partitioned the frequency domain into 19 frequency windows with a width of 20Hz from 0Hz to 300Hz, and with a width of 50Hz from 300Hz to 500Hz (the Nyquist frequency). The windows were set to better reflect the low-frequency details. Within each frequency window, the mean spectral density was extracted. The spectral features were extracted because they proved to be effective in classifying different types of head impacts (e.g., CF, MMA, car crashes) \cite{zhan2018discrimination}, which manifested that they contained the information related to the impact types. Therefore, to get more accurate models for different types of head impacts, we included these spectral features in the modeling. It should be noted that our previous study \cite{zhan2021clustering} reported 352 temporal and spectral features but only 350 were extracted in this study, because there were two overlapped features with the previous 160-feature set (i.e., the maximum values of the magnitudes of $\omega(t)$, $\alpha(t)$).

To sum up, there were 510 temporal and spectral features extracted from four types of kinematics ($a(t)$, $\omega(t)$, $\alpha(t)$, $j(t)$) which were used in the modeling of MPS and MPSR for the entire brain. Feature standardization was performed according to the following formula to avoid imbalanced weights for different features caused by feature range mismatches:

\begin{equation}
	\tilde{X}(i,j) = \frac{X(i,j)-\mathrm{mean}_i(X(i,j))}{\mathrm{std}_i(X(i,j))}
\end{equation}
	$\tilde{X}(i,j)$ denotes the standardized $j^{\text{th}}$ feature of the $i^{\text{th}}$ impact. $\mathrm{mean}_i()$ and $\mathrm{std}_i()$ denotes the mean and standard deviation over the variable $i$ based on the raw feature $X$. The parameters used in this study were calculated based on the training data in the hyperparameter tuning process and on the union of the training data and validation data in the model evaluation process (Please refer to details in Section 2E).
	
The feature extraction and signal processing were done in MATLAB R2020a (Austin, TX, USA). The data standardization was done in Python 3.7 with scikit-learn packages \cite{pedregosa2011scikit}. 

\subsection{Development of the basis deep learning model}
Based on the dataset HM, two separate basis MLHMs were developed to model the mapping from the kinematic features to the MPS and MPSR given by the FE simulations. To develop the models, the entire dataset HM (12,780 impacts) was randomly partitioned into 70\% training data for model training, 15\% validation data for hyperparameter tuning and 15\% test data for model performance evaluation, which is illustrated in Fig. 2A.

\begin{figure*}[htbp]
    \centering
    \includegraphics[width=0.95\linewidth]{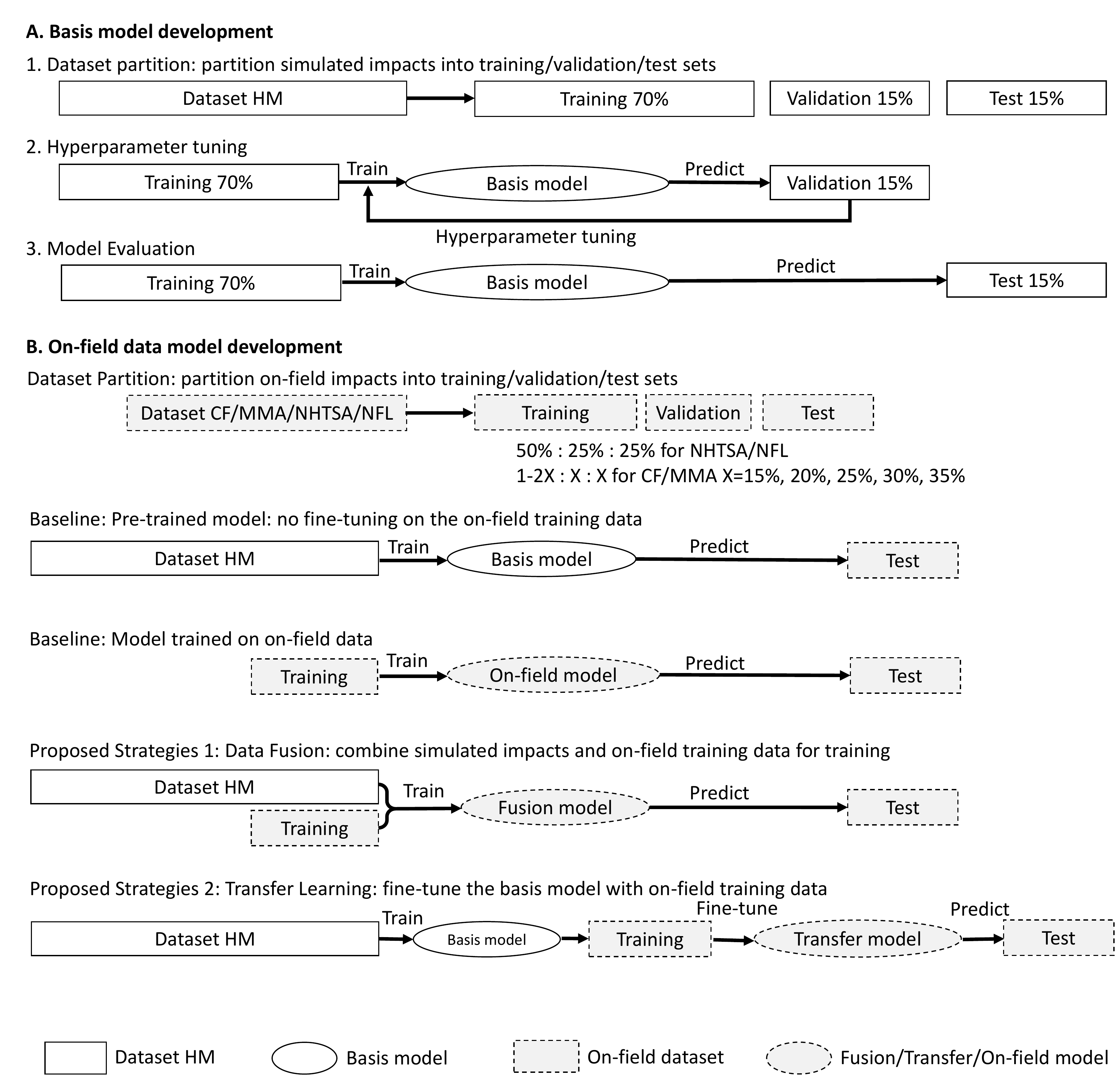}
    \caption{The illustration of the model development and assessment processes. The processes to develop the basis models developed for the dataset HM (A), and for the on-field datasets (B). The two proposed strategies in adapting for different on-field impacts and the baseline models we compared against (B). The datasets and models related to the on-field impacts are in dotted gray boxes, while those related to the simulated impacts (dataset HM and basis models) are in solid white boxes. The models are represented with oval boxes and the datasets are represented with rectangular boxes.}
    \label{pipeline}
    \vspace{-5mm}
\end{figure*}

The models shared the same network structure, which consisted of five layers (excluding the input layer (510 neurons) and the output layer (4124 neurons)): 1) hidden layer 1: 500 neurons coupled with the rectified linear unit (ReLU) as the activation function; 2) dropout layer 1 with 0.5 as the dropout rate; 3) hidden layer 2: 300 neurons coupled with the ReLU as the activation function; 4) dropout layer 2 with 0.5 as the dropout rate; 5) hidden layer 3: 100 neurons coupled with the ReLU as the activation function. The design of this structure followed the guidance of firstly condensing kinematics information from the 510 features and then predicting the output variable. The connection weights of the MLHMs were randomly initialized to break the model symmetry at the start of model training. To improve the model generalizability, the dropout and L2 penalty were used as forms of regularization. To drive the gradient descent and model training, the mean squared error (MSE) was used as the loss function. To boost the training efficiency and get faster convergence, the adaptive moment estimation (Adam) \cite{kingma2014adam} optimizer was used as the optimizer with 128 set as the batch size.

The hyperparameters in the development of the basis model included: the specific numbers of hidden neurons, the training epochs, the strength of L2 penalty, and the learning rate. They were tuned based on the prediction performance on 15\% validation data (prediction performance metrics shown in Section 2E).

Before training the models, we adopted the same data augmentation strategies of adding moderate Gaussian noise into the features which were used in our previous studies \cite{zhan2021rapid,liu2021boost,zhan2019feature} and proved to effectively improve model accuracy:

\begin{equation}
    \label{noise}
    Noise =c \times \mathrm{std}_i(X(i,j)) \times \delta, \quad \delta \sim N(0,1)
    \end{equation}
 where $j$ denotes the $j$-th feature and $c$ is the coefficient of noise level taking one of the values in ($0.01, 0.02$). Upon completing the data augmentation, the total number of training impacts was tripled.

Additionally, to further optimize the accuracy, we applied the logarithmic transform on the labels (MPS, MPSR) in the development of the basis deep learning model to stabilize the residual variance and avoid negative MPS and MPSR predictions. To use this transform, in the training stage, the MPS and MPSR values were transformed by taking its logarithm. In the prediction stage, the predicted values were inverse-transformed as MPS and MPSR by exponentiation.

The data augmentation and deep learning were performed in Python 3.7, using the Keras package with the TensorFlow 2.0 backend \cite{abadi2016tensorflow}. 

\subsection{Data fusion and transfer learning strategies}
As shown in our previous studies \cite{zhan2021rapid, zhan2021relationship}, many brain injury risk estimation models show that accuracy is not guaranteed across different types of head impacts, which may be caused by 1) that the limited on-field impacts cannot properly determine the model parameters and 2) the distribution shifts from the model training data (source domain) to the test data (target domain). Although we can develop the basis model on the simulated dataset, researchers typically attach more significance to the model performance on the on-field impacts than that on the simulated data for potential real-world applications. Therefore, to improve the accuracy of MLHMs on different types of on-field head impacts, we developed two strategies to adapt the basis models trained on dataset HM to better adapt to the on-field impacts, which are illustrated in Fig. 2B:

\textbf{Data fusion strategy}: the on-field impacts (CF/MMA/NHTSA/NFL) were firstly partitioned into the training data, validation data, and test data (Fig. 2B). The training data from the on-field dataset were fused with the larger dataset HM in the development of MPS and MPSR models. In this way, the model can benefit from the large number of impacts from the dataset HM and learn the general relationship between kinematics features and the MPS and MPSR. To account for the potential feature distribution variation from dataset HM to the on-field data, we recalculated the mean and standard deviation for feature data standardization based on the union of dataset HM and the on-field training data. We refer to this approach as data fusion because we combine the computer-simulated impacts and on-field impacts measured by sensors for model development.

Furthermore, to account for the potentially different MPS and MPSR distributions, for this strategy, in addition to the logarithmic transform, we adopted the data whitening approach according to the following formula:

\begin{equation}
	\tilde{Y}(i,j) = \frac{Y(i,j)-\mathrm{mean}_i(Y(i,j))}{\mathrm{std}_i(Y(i,j))}
\end{equation}

$Y(i,j)$ denotes the logarithmic-transformed MPS and MPSR of the $j^{\text{th}}$ brain element for the $i^{\text{th}}$ impact. The $\mathrm{mean}_i(Y(i,j))$ and $\mathrm{std}_i(Y(i,j))$ for each of the 4124 brain elements were calculated on the combination of dataset HM and training set of on-field impacts in hyperparameter tuning process, and on dataset HM, the training and validation sets of on-field impacts in model evaluation. $\mathrm{mean}_i(Y(i,j))$ and $\mathrm{std}_i(Y(i,j))$ were recorded to inverse-transform the prediction of the logarithmic-transformed MPS and MPSR. The assumption was that the dataset used to calculate $\mathrm{mean}_i(Y(i,j))$ and $\mathrm{std}_i(Y(i,j))$ represented a general distribution and in this way, the prediction model can also benefit from the overall distribution information. The logarithmic-transform was performed before the data whitening. In the prediction stage, the inverse-transformed MPS and MPSR need exponentiation to get the MPS and MPSR predictions.

\textbf{Transfer learning strategy}: upon partitioning the on-field impacts into training, validation, and test sets, we used the training set to fine-tune the connection weights of the basis models pre-trained on the dataset HM (Fig. 2B). The fine-tuning process was done by freezing the connection weights of the first 0/1/2 hidden layers and using the weights of the basis model as the new initialization before another around of training was performed on the on-field training impacts. The number of frozen layers, the number of training epochs, the learning rate were set as the hyperparameters, which were tuned on the on-field validation impacts. As the adapted models were derived from the basis models, we fine-tuned the models with fewer epochs and/or lower learning rate which led to model convergence without overfitting to the limited on-field training impacts. To address the potential different feature distribution, data standardization was performed based on the union of dataset HM and the on-field training impacts. Data augmentation mentioned in the previous section was also used on the on-field training impacts before the second round of training (fine-tuning) was performed.

\subsection{Model evaluation}
%Baseline: A (HM->), B (logstd), C(log), D(HM+on-field -> on-field)
%Metrics
%Task: MPS
%MPSR
%The varying ratio of train/test split [real-world small datasets: NHTSA, NASCAR]
To quantify the model accuracy of the series of MLHMs we developed on various datasets (HM/CF/MMA/NHTSA/NFL), the following metrics were used:

1) Mean absolute error (MAE): the MAE between the predicted MPS/MPSR and the reference MPS/MPSR given by the KTH model, firstly averaged over 4,124 brain elements and then averaged over all test impacts (calculation details shown in \cite{zhan2021rapid}). Finally, the summary statistics (mean, median, STD) were calculated over 20 parallel experiments with random dataset partitions. This metric was used because it has the same unit as the MPS/MPSR for researchers to easily perceive the model accuracy and compare the error with the injury threshold. 

2) Root mean squared error (RMSE): the RMSE between the predicted MPS/MPSR and the reference values given by the KTH model, which was calculated in a similar manner as the MAE \cite{zhan2021rapid}. This metric was used in addition to the MAE because it emphasizes the large errors more. It is also in the same unit as the MPS/MPSR for easy comprehension.

3) Coefficient of determination ($R^2$): the $R^2$ was firstly calculated between the predicted MPS/MPSR and the reference values on 4,124 brain elements and then averaged across the test impacts. The details have been shown in \cite{zhan2021rapid}. This metric was used because the majority of previous publications related to brain strain estimation models demonstrate model accuracy in this metric. In addition to the MAE and RMSE which directly reflect the accuracy in the same units as MPS/MPSR, the $R^2$ values further reflect the goodness of fit and how much variance in the reference values can be explained by the predicted values.

To evaluate the performance improvement of the data fusion strategy and the transfer learning strategy on the on-field datasets (CF/MMA/NHTSA/NFL), we compared them with the following four baselines, which were also described in Fig. 2B and Table 2:

1) The pre-trained model: the basis MLHMs trained on the dataset HM and then directly used to predict the MPS/MPSR of the on-field test impacts. This baseline model was used to show the effect of the fine-tuning (the second round of MLHM weight training) on the on-field training data.

2-3) The models trained on the on-field training set: the MLHMs were trained on the training set partitioned from the on-field datasets (CF/MMA/NHTSA/NFL) either with both the logarithmic transform and the data whitening on the labels, or with the logarithmic transform only. These two baseline models were used to directly show the effect of the transfer learning strategy and data fusion strategy: whether these two strategies enabled accurate predictions that may not be achieved by merely training on the limited quantity of the on-field data. These models were labeled in the format of CF 1 (with both logarithmic transform and data whitening) and CF 2 (with only logarithmic transform) in the result sections.

4) The data fusion model without data whitening: the MLHMs were trained on the union of the dataset HM and the training set partitioned from the on-field datasets (CF/MMA/NHTSA/NFL). This baseline model was used to show the effect of the data whitening used in the data fusion strategy. This baseline model was labeled as "Fusion 2" in the results to be distinct from the proposed data fusion strategy ("Fusion 1").

Furthermore, to evaluate the effect of the quantity of on-field training data on the model accuracy, we gradually varied the ratio of the on-field training, validation, and test data on two large on-field datasets (CF, MMA), which is referred to as "1-2X" (percentage of training data) and "X" (percentage of validation data and test data) in Fig. 2B. Then, under various scenarios with different quantities of on-field training data, the model accuracy with the data fusion strategy and the transfer learning strategy was compared with that from the baseline models. With this varying ratio of training, validation, and test data, we simulated the real-world application scenarios where users may not have large quantities of on-field data to train and fine-tune a MLHM to accurately predict MPS and MPSR. This experiment setting tested whether the proposed two strategies can achieve high accuracy with different extents of training data inadequacy.

It is important to note that the varying ratio experiment simulated different levels of on-field training data inadequacy. Besides, two smaller on-field datasets were also used as the validation of the model performances in real-world application scenarios when there is not a large quantity of on-field data for model training and fine-tuning. We included these two datasets because the performance of the two proposed modeling strategies could manifest whether the hardship of accurately predicting brain strain and strain rate can be solved with the two proposed strategies. Due to the limited quantity of data in these two datasets, we partitioned the datasets into 50\% training data, 25\% validation data and 25\% test data so that there were more than 10 validation and test impacts in each experiment to give reasonable and robust results.

\subsection{Statistical tests}
To ensure robust results and test statistical significance, we did 20 parallel experiments for each prediction task and used the Wilcoxon signed-rank test to compare the accuracy (in MAE) of the models developed on the two proposed strategies and that of the baseline models. The paired t-test was not used because the Shapiro-Wilk test rejected the normal distribution assumption on some of the results; the Wilcoxon signed-rank test do not rely on the normality. The most conservative p-values were reported in the result sections, which held true for all the pairwise comparisons (e.g., between the transfer learning model and each of any other model), unless reported otherwise in more details. 

\section{Results}
\subsection{Model performance on dataset HM}
The abbreviations for model reference used in the result sections are shown in Table 2.
\begin{table*}
		\centering
		\caption{The abbreviations for model reference.}
		\label{tab:2}       % Give a unique label
		\begin{tabular}{cc}
			\hline\noalign{\smallskip}
			Abbreviation & Model \\
			\noalign{\smallskip}\hline\noalign{\smallskip}
			Transfer & The MLHM with transfer learning strategy (pre-trained on dataset HM and fine-tuned on on-field training set) \\
			Pre-train & The baseline MLHM trained on the dataset HM and directly used to predict the test impacts \\
			CF/MMA/NFL/NHTSA 1 & The baseline MLHM trained on the on-field training data with logarithmic transform and data whitening \\
			CF/MMA/NFL/NHTSA 2 & The baseline MLHM trained on the on-field training data with logarithmic transform only\\
			Fusion 1 & The MLHM developed with the data fusion strategy with logarithmic transform and data whitening \\
			Fusion 2 & The baseline MLHM developed with the data fusion strategy with logarithmic transform only \\
			\hline\noalign{\smallskip}
		\end{tabular}
	\end{table*}
	
According to the process of developing the basis MPS/MPSR models introduced in Section 2C and the metrics introduced in Section 2E, the basis models were built and evaluated on the dataset HM. The results are shown in Table 3. The model trained on 70\% of the dataset HM (8,946 training impacts in each experiment) showed the ability to accurately predict MPS and MPSR of the entire brain for the test impacts (1,917 test impacts in each experiment), with an averaged MAE of 0.015 for MPS prediction and 2.818 $s^{-1}$ for MPSR prediction (averaged over twenty paralleled experiments with random dataset partitions). The accuracy for MPS and MPSR prediction can also be reflected by the mean $R^2$ values: 0.860 and 0.786, respectively, which indicates that the basis MLHMs are accurate in predicting whole-brain MPS and MPSR. The MAE for the MPS prediction was much smaller than the differences of brain strain seen in injury and non-injury cases (between 0.3 and 0.4) \cite{ho2007dynamic,kleiven2007predictors}. The MAE for the MPSR prediction was also much smaller than the concussion threshold (around 25$s^{-1}$) \cite{patton2015biomechanical}, the threshold for accurate axonal injury prediction in large animal models (120$s^{-1}$) \cite{hajiaghamemar2020embedded} and the threshold for accurate brain contusion volume prediction in small animal models (2500$s^{-1}$ \cite{donat2021biomechanics}).

\begin{table*}
		\centering
		\caption{The hyperparameters and the model performance metrics of the basis MPS and MPSR models developed on dataset HM.}
		\label{tab:2}       % Give a unique label
		\begin{tabular}{ccccccccc}
			\hline\noalign{\smallskip}
			Output & Learning Rate & Training Epochs & L2 Regularization & MAE (Mean $\pm$ STD) & RMSE (Mean $\pm$ STD) & $R^2$ (Mean $\pm$ STD) \\
			\noalign{\smallskip}\hline\noalign{\smallskip}
			MPS & 0.0003 & 3000 & 0.01 & 0.015 $\pm$ 0.001 & 0.024 $\pm$ 0.001 & 0.860 $\pm$ 0.009\\
			MPSR & 0.0005 & 3000 & 0.05 & 2.818 $\pm$ 0.080 & 7.145 $\pm$ 1.294 & 0.786 $\pm$ 0.038 \\
			\hline\noalign{\smallskip}
		\end{tabular}
	\end{table*}

\subsection{Model performance on large on-field datasets}
Although high accuracy is achieved on the dataset HM, we attach significance to the model performance on the on-field datasets for sake of TBI risk monitoring. Therefore, two strategies, data fusion and transfer learning, were proposed and compared with the baseline models. The model accuracy (in MAE) on the two larger on-field datasets (CF and MMA) with varying ratios of training, validation, and test data were shown in Fig. 3 (RMSE and $R^2$ results shown in Fig. S1 and S2).

According to the results on dataset CF for the whole-brain MPS prediction, the transfer learning models were the most accurate ($p<0.01$), with MAE smaller than 0.015, except that with 30\% on-field training impacts (91 impacts), there was no statistically significant difference in MAE between the transfer learning model and the data fusion model (Fusion 1) ($p>0.05$). The data fusion models with both logarithmic transform and data whitening (Fusion 1) did not show significantly better performance when compared with the model trained merely on the CF training data ($p>0.05$), except that with 30\% on-field training impacts, the Fusion 1 model was significantly more accurate than this baseline model ($p<0.001$). For the whole-brain MPSR prediction, the transfer learning models were also the most accurate models ($p<0.01$), with MAE smaller than 3 $s^{-1}$ under all training set ratios, while the data fusion models (Fusion 1) were the second most accurate and significantly better than the baseline models ($p<0.001$).

On dataset MMA, for the MPS prediction, the transfer learning models were the most accurate under five different ratios of training data ($p<0.001$), with the MAE smaller than 0.015. The data fusion models (Fusion 1) did not significantly improve the MPS prediction accuracy when compared with the models trained only on the MMA training data ($p>0.05$). For the MPSR prediction, the transfer learning models were the most accurate models under varying training set ratios ($p<0.01$). The data fusion models (Fusion 1) were generally the second most accurate models, but did not show significant improvement over the baseline model (MMA 1) on every training set ratio. These results indicate that the pre-training based on the dataset HM and fine-tuning based on the dataset CF/MMA can adapt the models to be more accurate for the MPS and MPSR estimation of these two specific types of on-field data. On the contrary, though using the dataset HM as well, the data-fusion-based MLHMs (Fusion 1) generally lag behind the transfer learning strategy in accuracy. The models may also improve the model accuracy to a limited extent, but the improvement is not guaranteed. Compared with the MPS MLHMs we previously developed \cite{zhan2021rapid} with 70\% MMA training data, the transfer learning MLHMs reached a mean MAE around 0.013 and a mean $R^2$ around 0.7, which is evidently more accurate than the previously developed MLHMs (mean MAE around 0.025, mean $R^2$ around 0.6).

As for the comparison between the transfer learning models and the pre-trained models, the former models were significantly more accurate than the latter models ($p<0.01$). This indicates that the additional model fine-tuning (the second round of MLHM connection weight training on the on-field training data) improved the model for specific on-field prediction tasks. When the two data fusion models (Fusion 1/2) were compared, it is shown that the data whitening process contributes to the model accuracy improvement, as the Fusion 1 models were significantly more accurate than the Fusion 2 models. Furthermore, the MPS model trained only on the on-field training data tended to be more accurate with more training data, which indicates that one limitation of the accuracy of models merely based on the on-field data is the quantity of the training impacts.

\begin{figure*}[htbp]
    \centering
    \includegraphics[width=0.95\linewidth]{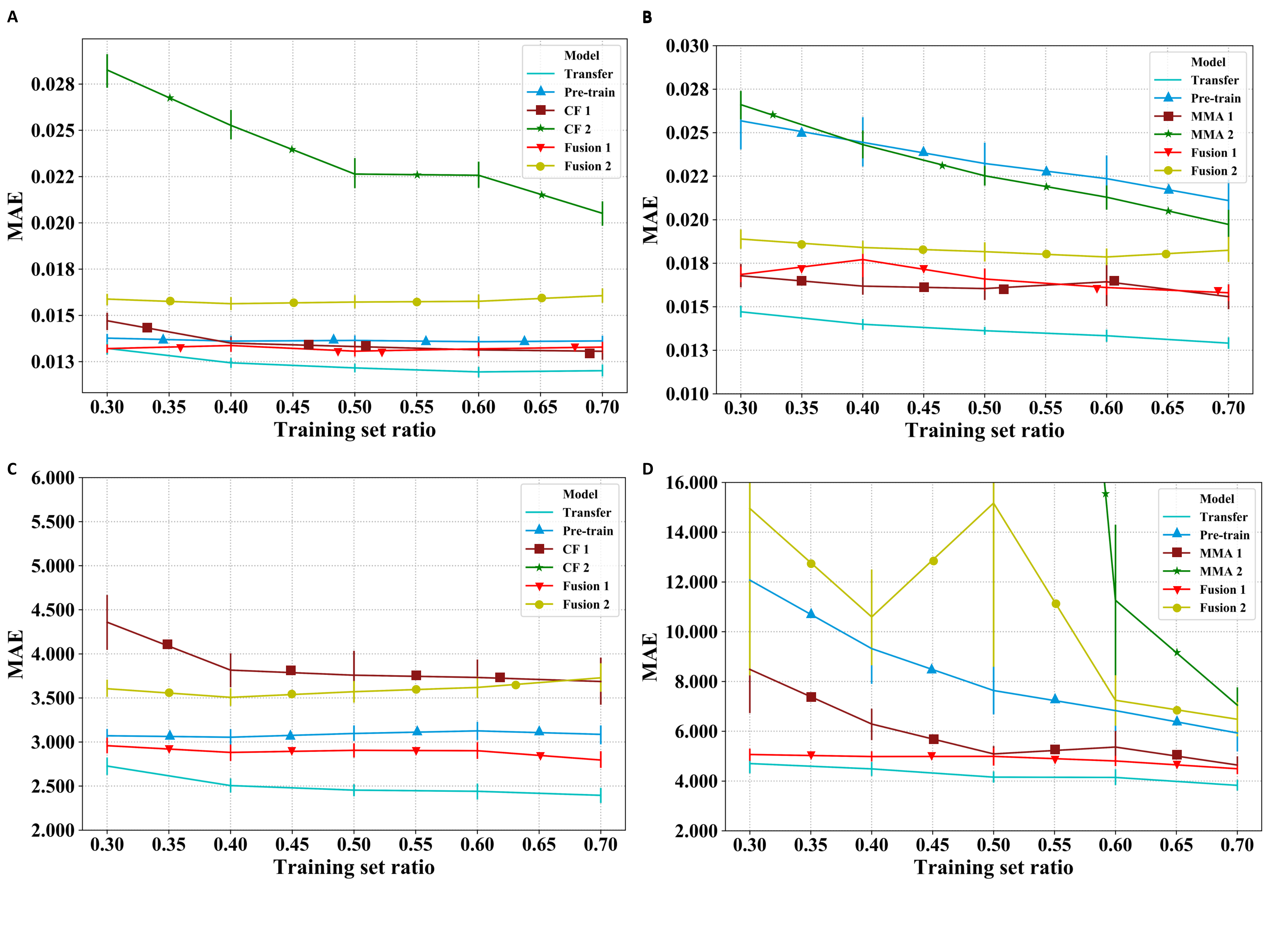}
    \caption{The model accuracy of predicting MPS and MPSR on the dataset CF and dataset MMA with different ratios of training data. The mean absolute error (MAE) of the MPS prediction models on dataset CF (A) and dataset MMA (B). The MAE of the MPSR prediction models on dataset CF (C) and dataset MMA (D). Due to the huge variation among different models, the MAE ranges were selected to clearly show the most accurate models. Note that the training set ratio refer to the percentage of on-field data used for training purposes, which is the "1-2X" in Fig. \ref{pipeline} and in Section 2E.}
    \label{mae_large}
    \vspace{-5mm}
\end{figure*}

\begin{figure}[htbp]
    \centering
    \includegraphics[width=\linewidth]{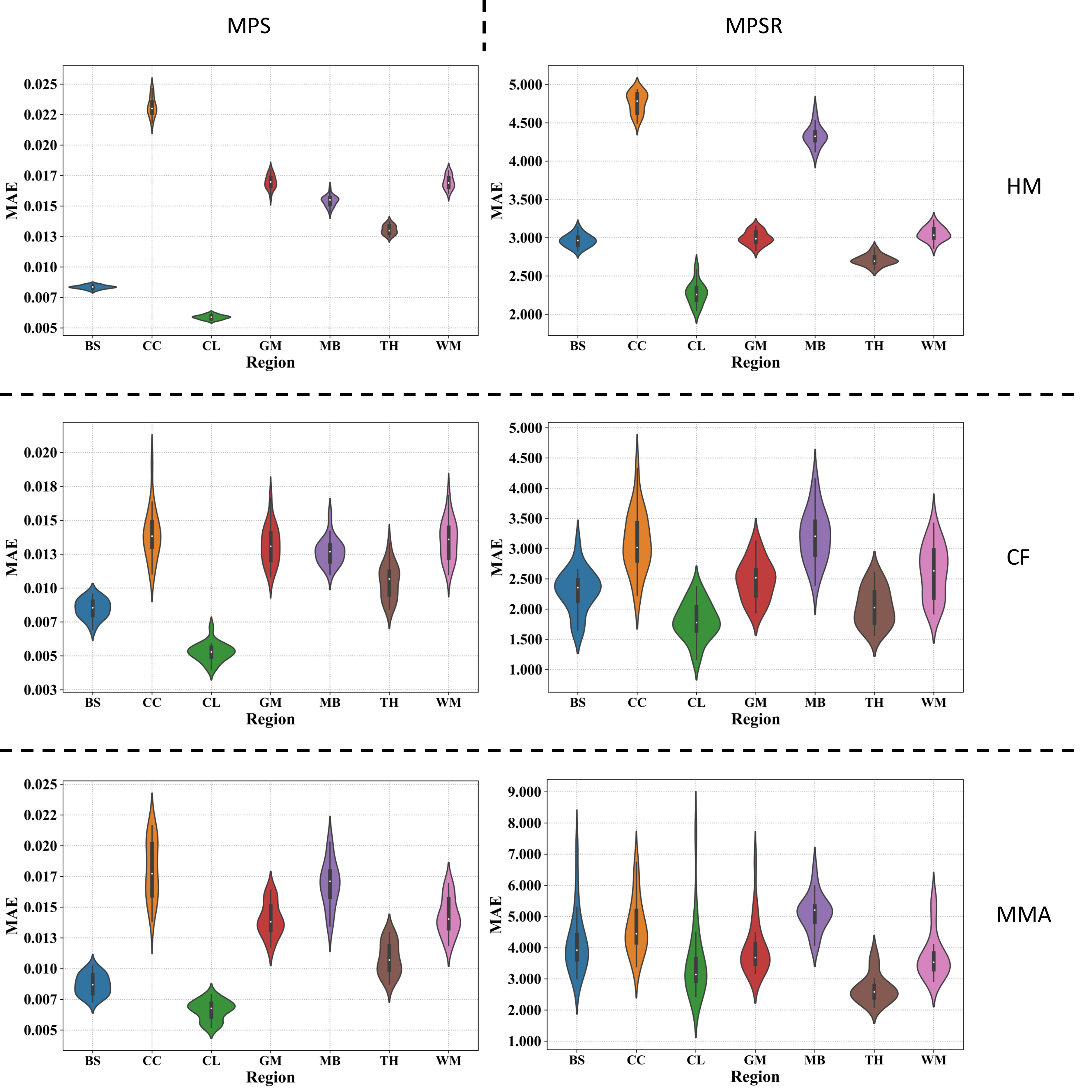}
    \caption{The region-specific mean absolute error of the transfer learning models on dataset HM, dataset CF and dataset MMA with 70\% training data and 15\% test data (mean and STD over 20 parallel experiments). BS: brainstem, CC: corpus callosum, CL: cerebellum, GM: gray matter, MB: midbrain, TH: thalamus, WM: white matter.}
    \label{region}
    \vspace{-5mm}
\end{figure}

In addition to showing the accuracy of the transfer-learning-based models over all brain elements, we have also calculated the MAE on the brain elements based on different brain regions and the results are shown in Fig. \ref{region} and Table S2. The results show that overall, the MPS/MPSR models based on transfer learning manifest high accuracy on various brain regions, with a mean MAE smaller than 0.025 for MPS prediction and a mean MAE smaller than 5.2 $s^{-1}$ for MPSR prediction across different brain regions. The MLHMs tended to be slightly more accurate on some regions (brainstem, cerebellum, thalamus) but slightly less accurate on other regions (corpus callosum, midbrain), in terms of MAE. The error was well below the injury thresholds in MPS and MPSR which are mentioned previously \cite{ho2007dynamic,kleiven2007predictors,hajiaghamemar2020embedded,patton2015biomechanical,donat2021biomechanics}.

\subsection{Model performance on small on-field datasets}
\begin{figure}[htbp]
    \centering
    \includegraphics[width=0.95\linewidth]{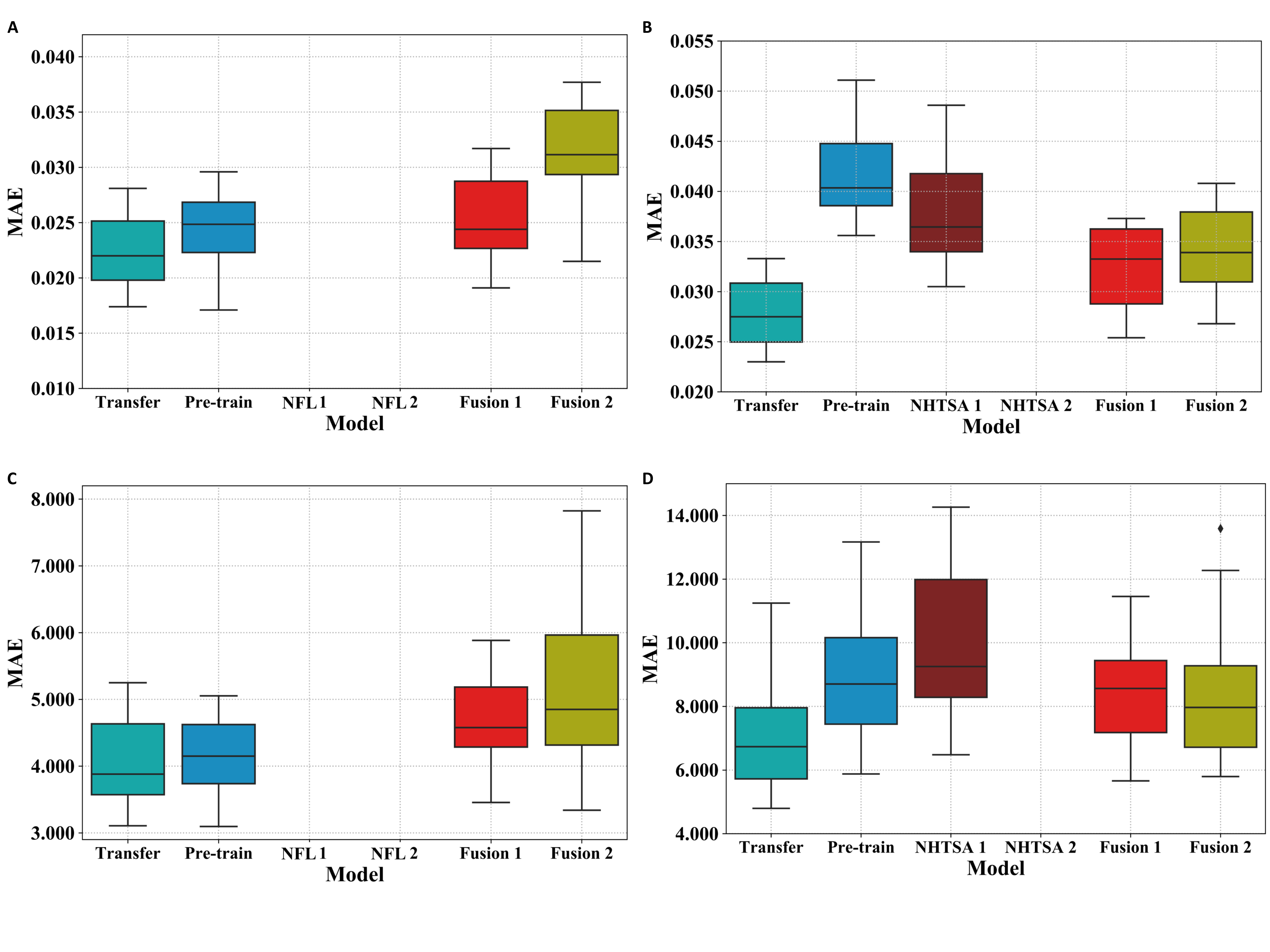}
    \caption{The model accuracy of predicting MPS and MPSR on the dataset NFL and dataset NHTSA with 50\% training data and 25\% test data for each dataset. The mean absolute error (MAE) of the MPS prediction models on dataset NFL (A) and dataset NHTSA (B). The prediction MAE of the MPSR prediction models on dataset NFL (C) and dataset NHTSA (D). Due to the large variation among different models, the MAE ranges were selected to clearly show the most accurate models. Note that the models trained on NFL dataset only failed to predict the MPS and MPSR for the test impacts due to numerical errors.}
    \label{mae_small}
    \vspace{-5mm}
\end{figure}

The results of the model accuracy were further validated on the two real-world on-field datasets (NFL: 36 impacts, NHTSA: 48 impacts) with far fewer impacts. On these two small datasets, the percentage of training, validation, and test data were set to be 50\%, 25\% and 25\%, to ensure enough test impacts in each experiment for robust accurate estimates. The MPS and MPSR model accuracy in MAE is shown in Fig. 4 (RMSE and $R^2$ results shown in Fig. S3 and S4). On dataset NFL, the median MAE based on the transfer learning MPS model was smaller than 0.025 and the median MAE based on the transfer learning MPSR prediction was smaller than 4 $s^{-1}$. For the MPS prediction, the transfer learning model was the most accurate ($p<0.001$). For MPSR prediction, the transfer learning model was more accurate than the data fusion models ($p<0.001$) but did not show statistical significance with the pre-trained model ($p>0.05$). On dataset NHTSA, the transfer learning models were the most accurate models for both MPS prediction and MPSR prediction ($p<0.001$), with a median MPS prediction MAE smaller than 0.028 and a median MPSR prediction MAE smaller than 7 $s^{-1}$. The two models trained on NFL training data (18 impacts) failed to predict the MPS and MPSR due to the fact that some predictions caused numerical overflow on 32-bit floating point, and therefore we deemed these two models unable to predict MPS and MPSR on dataset NFL (Fig. \ref{mae_small}). In addition, the data fusion strategies were also effective in accurately predicting the MPS and MPSR, but generally, the models directly trained on the limited NFL/NHTSA data failed to accurately predict the MPS and MPSR. These results indicate that both of the proposed strategies were effective in achieving accurate MPS and MPSR predictions which otherwise may not be achieved by directly training on the limited on-field data, while the transfer learning models were still the best performing models on all cases.

\subsection{Visualization of MPS and MPSR predictions}
\begin{figure}[htbp]
    \centering
    \includegraphics[width=0.99\linewidth]{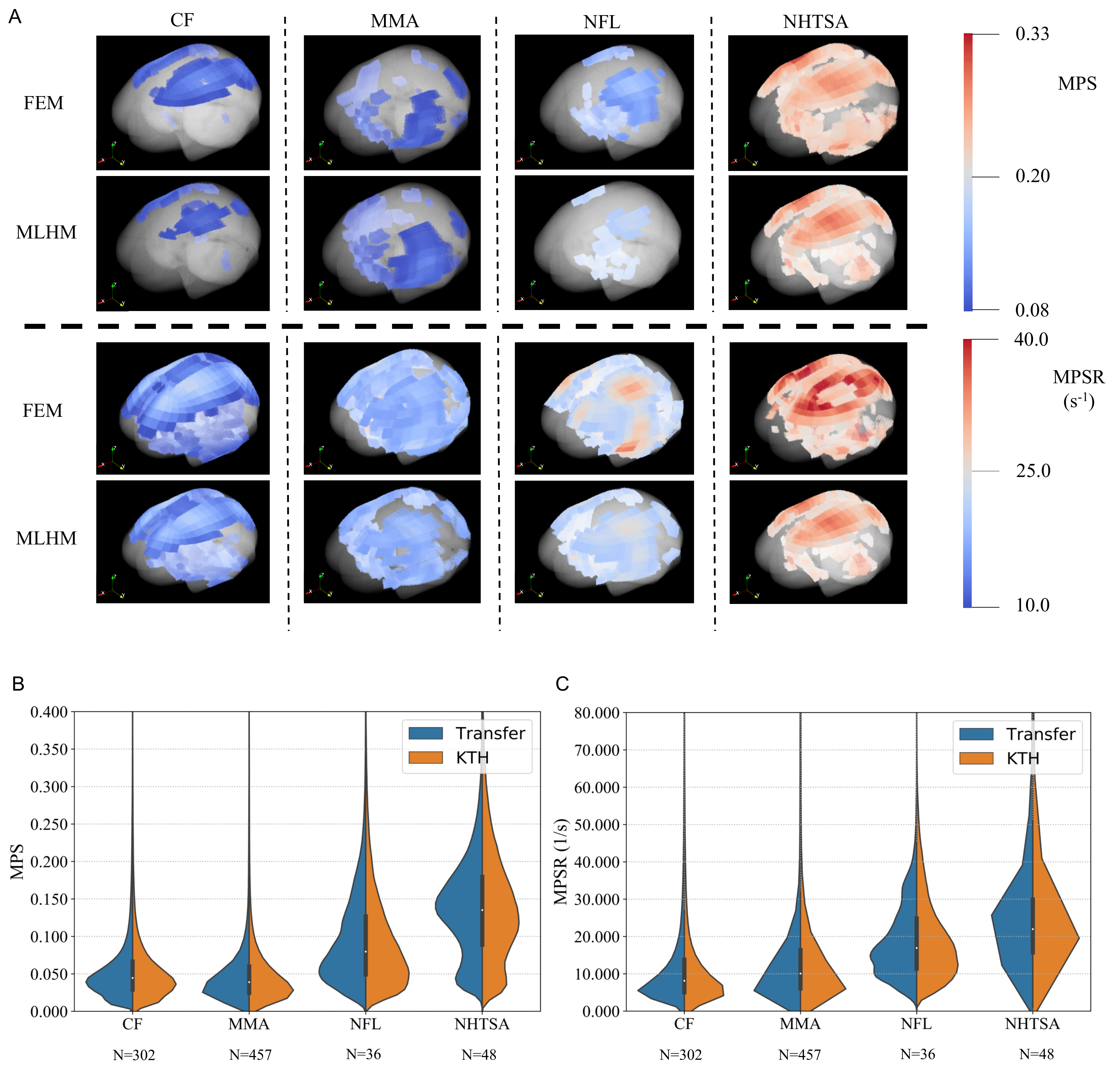}
    \caption{The visualization of the accuracy of MPS/MPSR prediction with example point clouds and overall distribution violin plots. The point clouds predicted and reference MPS/MPSR of the example impacts based on the transfer learning models (A), and the violin plot of the overall distribution of predicted and reference MPS/MPSR for the on-field datasets (B,C). The example impacts were selected based on the median 95th percentile MPS/MPSR for each dataset. The high-strain and high-strain-rate regions were colored to visualize the accuracy of the predictions given by the transfer-learning models.}
    \label{3D}
    \vspace{-5mm}
\end{figure}

To visually examine the model accuracy in predicting MPS and MPSR. We selected the example test impacts from dataset HM, CF, MMA, NFL and NHTSA, by firstly ranking the predictions with 95th percentile MPS and MPSR and taking the the median, and plotted the 3D brain MPS and MPSR maps given by the model predictions (Fig. \ref{3D} A). Note that 95th percentile rather than the 100th percentile was selected to avoid potential computational artifacts from analysis \cite{panzer2012development}. Here the predictions given by transfer-learning-based MLHMs and the reference KTH model (FEM) output were compared with 3D point clouds. It can be shown from the results that, for each type of on-field impacts, the high-strain and high-strain-rate regions predicted by the MLHMs were similar to those output by the FEM. Besides the accuracy shown in the example 3D visualizations, we plotted the overall distributions of the predicted MPS/MPSR on the on-field datasets and compared them against the reference values given by the KTH model (Fig. \ref{3D} B and C). It can be seen that the transfer-learning-based MLHMs predict the overall distribution of the MPS and MPSR accurately. Particularly, on the two small on-field datasets (NFL and NHTSA), with limited impacts for fine-tuning, the transfer-learning MLHMs can reproduce the distribution well, which indicates that the MLHMs can be applied to monitoring MPS and MPSR caused by head impacts.. 

\section{Discussion}
%1. What has been done in this study
In this study, we proposed two strategies, data fusion and transfer learning, to optimize the machine learning head models (MLHMs) to more accurately predict the brain strain (MPS) and strain rate (MPSR) for different types of on-field impacts. This study aims to address the problem of the large computational cost associated with calculation of brain strain and strain rate with FEM. Further, this study sought to address the limitations of previously developed MLHMs: 1) they are only able to predict brain strain and 2) their accuracy are limited by the quantity of the training data. As a result, for the on-field impacts collected from college football, MMA, car crashes, and NFL, the series of MPS/MPSR models developed with the transfer learning strategy (when we pre-train the MPS/MPSR models on the large dataset of head model simulated impacts and fine-tune them with small quantities of on-field impacts), generally turn out to be the most accurate models when compared with the models merely trained on the on-field data and the models trained by fusing the large dataset of head model simulated impacts and the on-field impacts. The results indicate that the transfer learning strategy enables the learned general brain dynamics, i.e., the relationship between head kinematics and MPS/MPSR, to be transferred to on-field impacts for more accurate MPS/MPSR estimation.

%2. Contribution as MLHM: fast accurate supplant of FEM, high dimenstionality against BIC, application-wise contribution
This study provides two significant contributions to the field of brain injury research. First, the developed series of MLHMs significantly reduce the computational time cost resulting from conventional FEM. The series of MLHMs, once trained, can estimate the MPS and MPSR of an impact within 1 ms with a personal computer (8GB RAM, Intel Core i5-6300U CPU). On the contrary, it takes hours for the KTH model to model an impact and output its MPS and MPSR with a more powerful computer (16GB RAM, Intel Core i7-6800K CPU), which makes it almost impossible to output real-time warnings for early intervention approaches to protect contact sport athletes soon after they sustain an impact. This significant reduction of computational time, without loss of accuracy in MPS and MPSR estimation, enables researchers and clinicians to efficiently perform TBI risk estimation in real-time. Therefore, this series of MLHMs can be applied by sport team managers and clinicians to quickly monitor the brain MPS and MPSR for athletes, on the premise that the impact kinematics are measured by wearable devices \cite{greenwald2008head,camarillo2013instrumented}. Previous evidence suggests that high brain MPS and MPSR correlate with pathological changes related to TBI, including  traumatic axonal injury (TAI) and blood-brain-barrier (BBB) disruption \cite{o2020dynamic,hajiaghamemar2020head,hajiaghamemar2020embedded}. Further, previous studies have also shown that early detection and intervention improve patient recovery from TBI \cite{guiza2017early}. Therefore, by combining wearable sensor technology with the series of MLHMs developed in this study, better precautionary and protective decisions for contact sport athletes can be made in a timely manner, such as resting and substitution suggestions, better protective gear suggestions, and even medical imaging instructions.

The second major contribution of this study is that the MLHMs developed in this study provide the users with the region-specific and high-dimensional brain strain and strain rate information for TBI diagnosis. Although various brain injury criteria have been developed in previous studies to address the time consumption of FEM \cite{takhounts2003development,takhounts2013development,gabler2018development,gabler2019development,zhan2021relationship}, these brain injury criteria only output one value to summarize the overall brain injury risks and are hard to directly interpret for users without an injury biomechanics background. They cannot output region-specific detailed brain strain/strain rate information. In contrast, the MLHMs are able to predict the brain strain and strain rate for every brain element and visualize them in a 3D map, which makes it easier for users to compute region-specific risk information and visually evaluate the TBI risks caused by head impacts without the need to fully understand the underlying biomechanics. 

%3. Novelties of this study: 1) the transfer learning: accurate and generalizable across different on-field datasets, to be more accurate on small on-field dataset which otherwise cannot be reached; 2) MPSR: further diversify the output information for the researchers from MPS
Furthermore, when compared with the MLHMs previously developed by our group and other researchers \cite{ghazi2021instantaneous,zhan2021rapid}, this study bears unique innovations. First, this study provides the users with novel MPSR prediction models which have not been developed by previous studies. Although the brain strain, particularly MPS, has been shown to correlate well with concussion and BBB disruption \cite{bain2000tissue, bar2016strain, donat2021biomechanics, cater2006temporal,fahlstedt2015correlation,o2020dynamic}, the predictive power of strain rate, particularly MPSR, is a very important complement to MPS.  Researchers have manifested that MPSR correlates well with TAI and therefore, TBI is not only dependent on the brain strain but also on the strain rate \cite{hajiaghamemar2020embedded,hajiaghamemar2020head}. In addition to the MPS prediction models, this study further enables fast diagnosis using MPSR. 

More importantly, this study leverages transfer learning and data fusion strategies to address inferior model performance on small on-field datasets. In our previous studies, the generalizability of TBI risk estimation models (including brain injury criteria and MLHMs) has been limited \cite{zhan2021rapid,zhan2021relationship}. Even though the models may perform well on the datasets where they are trained and developed, their accuracy may deteriorate when applied to specific types of on-field impacts. Particularly, this often occurs when they are applied to estimate the risks for impact types with inadequate training data \cite{zhan2021rapid}. Due to the limited on-field training data, MLHMs developed mainly based on simulated head impacts cannot accurately estimate both the general brain dynamics and the dataset-specific characteristics intrinsic in certain on-field impacts. Therefore, the MLHMs may overfit to the simulated head impacts, which leads to their inferior accuracy when applied to certain on-field impact types where the on-field data distribution is far different from that of the simulated impacts. With the transfer learning strategy and data fusion strategy, the MLHMs are more accurate on the smaller on-field datasets. Because researchers usually attach more significance to the model performance on on-field impacts, this study achieves the goal of further optimizing the on-field impact risk estimation accuracy for real-world applications.

%4. Result discussion 1: Y [influenced by the Y distribution already seen] and X->Y [brain dynamics can be transfered]; parallel structure/series structure; description of the results;
In this study, the two proposed strategies for better on-field impact MPS/MPSR predictions generally work well in reducing the prediction MAE and increasing prediction $R^2$. Although both strategies attempt to leverage the information gained from the large dataset of simulated head impacts, the transfer learning strategy generally outperformed the data fusion strategy on almost all the experiments in this study. This can be due to 1) the different information processing protocols of these two different strategies, and 2) how these two strategies leverage the prior knowledge unique to their methods for estimating MPS/MPSR. The data fusion strategy processes the information in parallel by directly combining the simulated impacts and the on-field data, while the transfer learning strategy processes the information in series by first modeling on the simulated impacts and then on the on-field impacts. The parallel structure generally weighs the simulated impacts and on-field impacts equally, while the series structure attend more to the on-field data because although the model parameters in the fine-tuning process are initialized based on the simulated data, the ultimate parameters of the MLHMs are determined by the on-field data. As a result, the transfer-learning-based models can better adapt to the on-field data distribution. 

These two different strategies leverage different prior knowledge learned from the training data. The data fusion strategy adopts the data whitening of the labels (MPS, MPSR), which incorporates the distribution information of the simulated data and on-field data. The distribution of MPS and MPSR of the training data (i.e., the label distribution) can be regarded as the prior knowledge that is leveraged in the prediction. Therefore, the predictions are influenced by the MPS and MPSR distributions in the training data via the data whitening process. 
On the contrary, the transfer learning strategy leverages the prior knowledge stored in the model parameters (MLHM connection weights), which reflects the predictor-label relationship, rather than the label distribution. With the transfer learning strategy, the MLHMs learns the brain dynamics, which are the intrinsic characteristics of brain, in the pre-training process. Then, the fine-tuning process enables the MLHMs to adapt the dataset-specific characteristics intrinsic to the on-field impacts. In other words, the relationship between the kinematics features and the brain strain/strain rate is transferred, while what is fine-tuned is the dataset-specific characteristics.
To sum up, in addition to the common part of learning the mapping from predictors (kinematics) to the labels (MPS and MPSR), which are shared by both strategies, the data fusion strategy further leveraged the prior knowledge in the label distributions via data whitening, while the transfer learning strategy further leveraged the prior knowledge in the predictor-label relationship via model parameter fine-tuning. 

%5. Result discussion 2: Drift description: closeness of the impacts (follow BIC)
The variations in model performance across types of on-field impacts indicate the difference between on-field data and the head model simulated impacts. For instance, the pre-trained models without fine-tuning show high accuracy on the dataset CF and dataset NFL in both MPS and MPSR predictions. There were even no statistically significant differences between the transfer learning model and the pre-trained model in MPSR prediction on dataset NFL. These results indicate that the dataset HM, dataset CF, and dataset NFL are generally similar in their data distributions. On the contrary, the pre-trained models show significantly inferior accuracy on dataset MMA in predicting MPS and MPSR, which are even less accurate than the models trained merely on the on-field MMA impacts. This observation indicates that the models trained on dataset HM do not generalize well onto MMA impacts and the differences between these two datasets are more evident than those among the HM, CF, and NFL datasets. This observation may be explained by the fact that the head model simulated impacts are generally simulated for the research of American football impacts, and therefore, the simulated impacts may better represent the characteristics shown in college football or NFL impacts \cite{zhan2021relationship}. Additionally, the MMA head impacts are measured without protective helmets while the football impacts are measured with protective helmets, which may effectively change the impact responses of the head \cite{vahid2021computational}. The distribution drifts reflect the different characteristics of various types of head impacts \cite{zhan2021classification} and according to the results, the transfer learning is effective in capturing the distribution drifts across head impact types and enables more accurate MPS and MPSR estimation. Furthermore, for the models trained only on the on-field training data, the accuracy generally increases with more training data. It is possible that with adequate on-field data, these models can approximate the best performing transfer-learning models.

Additionally, previous publications have focused on region-specific brain injuries caused by head impacts and region-specific injury management, where emphasis has been placed on brain regions such as the brainstem \cite{kochanek2019management}, corpus callosum \cite{hernandez2019lateral,reeves2005myelinated}, and cerebellum \cite{potts2009models,park2007cerebellar}. As we investigated the model accuracy across different brain regions, the models turned out to be generally accurate across different brain regions, with slightly higher accuracy on the brainstem, cerebellum, and thalamus than the corpus callosum in terms of MAE. Because one of the major advantages of MLHMs over brain injury criteria is the whole-brain strain and strain rate predictability, with the MLHMs, researchers can extract the region-specific brain strain and strain rate information to evaluate TBI risk information for different brain regions. With prior knowledge of human physiology and the region-specific strain and strain rate information, researchers and clinicians can focus more on the potential outcomes that may be related to region-specific pathology including BBB disruption and TAI.

%6. Limitations of this study
Although this study shows the effectiveness of data fusion and transfer learning in optimizing TBI risk estimation, there are several limitations that need to be pointed out. First, we used the KTH model as the reference method to determine the accuracy of the MLHMs. The KTH model is limited when we compare it with the state-of-the-art finite element head models (FEHM) \cite{li2021anatomically,fahlstedt2021ranking}. For instance, the gyri and sulci can have significant influences on the behavior of FEHM. However, they are not included in the KTH model. Future studies can focus on the development of MLHMs based on more advanced FEHM. 

Second, the brain difference and head shape variation among different individuals are not considered in this study. There have been previous studies showing that individual differences can bear significant influence on brain strain \cite{li2021anatomically}. Therefore, to make the MLHMs even more accurate across different people, future studies can take other personal predictors into the models.

Furthermore, the series of MLHMs developed in this study are based on the features extracted from the kinematics (the linear acceleration, the angular velocity, the angular acceleration, and the angular jerk). Therefore, the quality of the engineered features set an upper threshold of model performances. We adopted this feature engineering approach because: 1) these features have been shown to be very effective in aggregating the temporal and spectral information from the signals and predicting brain strain \cite{zhan2021rapid,zhan2021predictive}; 2) there are limited quantities of on-field data to train the end-to-end MLHMs out of the raw signals; and 3) due to the different triggering patterns, sampling rates, method of attachment to the head, recording time window lengths of various measurement devices to collect the on-field data and the different patterns of impact kinematics across various impact types, there may be issues such as the time mismatch of the start/peaks of the impacts collected from different devices. Therefore, with limited training data, directly feeding the signals into MLHMs may not lead to effective feature extraction. In contrast, the feature engineering approaches shown in this study can be effective to summarize the information stored in the signals which we thought could contribute to brain strain and strain rate without misleading the MLHMs with temporal details. However, as more data are available, more data-driven feature extraction approaches such as the convolutional neural network (CNN), variational autoendocer (VAE), and recurrent neural network (RNN) with long short term memory (LSTM) can be tested and applied to get more accurate models without the feature engineering step. These types of complex neural network structures are effective in extracting information with large quantities of training data without human knowledge and feature designs, and therefore may be able to further optimize the MLHMs.

\section{Conclusion}
This study proposes data fusion and transfer learning strategies to develop a series of optimized machine learning head models (MLHMs) to more accurately predict the brain strain (MPS) and strain rate (MPSR) for different types of on-field impacts. The results show that the MLHMs developed with the transfer learning strategy are generally the most accurate to estimate the MPS and MPSR for on-field impacts collected from college football, MMA, car crashes, and NFL, when compared with models trained on the on-field data only and models trained on the union of the head model simulated impacts and the on-field impacts. The results indicate that the transfer learning strategy enables the MLHMs to learn the general brain dynamics from large quantities of simulated head impacts and then adapt to specific types of on-field impacts via fine-tuning. The MLHMs developed in this study enables fast and accurate computation of brain strain and strain rate resulting from various on-field impacts for real-time risk estimation of traumatic brain injury, which cannot be achieved with the conventional finite element modeling and previous MLHMs developed by researchers.

\section{Acknowledgement}
The authors express their gratitude for the suggestions and assistance in transfer learning given by Mr. Fanjin Wang from University College London.

\bibliographystyle{IEEEtran}
\bibliography{reference}

\end{document}